\pdfoutput=1
\documentclass[5p]{elsarticle}

\usepackage{amssymb}

\usepackage{hyperref}
\hypersetup{colorlinks,pdfstartview = FitH,bookmarksopen = true,bookmarksnumbered = true,linkcolor = black,plainpages = false,hypertexnames = false,citecolor = black}

\usepackage{amsmath}

\biboptions{comma,sort&compress}

\usepackage{caption}
\usepackage{enumitem}

\usepackage[absolute]{textpos}
\usepackage{calc}
\newcommand{\copyrightstatement}{
    \begin{textblock*}{\textwidth}(0.5in,\textheight+\footskip+1.25in)    
        \noindent
        \footnotesize
				\begin{minipage}{0.15\textwidth}
				\includegraphics[width=0.99\textwidth]{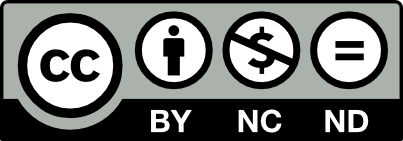}
				\end{minipage}
				\begin{minipage}{0.85\textwidth}
				 \url{https://dx.doi.org/10.1016/j.robot.2013.08.010} \\
				 \copyright {\the\year} This manuscript version is made available under the CC-BY-NC-ND 4.0 license \\
				 \url{http://creativecommons.org/licenses/by-nc-nd/4.0/}
				 \end{minipage}
    \end{textblock*}
}

\journal{Robotics and Autonomous Systems}

\begin{document}

\copyrightstatement


\begin{frontmatter}




\title{Optimal routing strategies for autonomous underwater vehicles in time-varying environment}


\author{Mike Eichhorn}
\ead{mike.eichhorn@tu-ilmenau.de, Mike\_Eichhorn@gmx.de, mike.eichhorn@ieee.org.}
\address{Institute for Ocean Technology, National Research Council Canada Arctic Avenue, P.O. Box 12093
St. John's, Newfoundland A1B 3T5, Canada}

\begin{abstract}
This paper presents a mission system and the therein implemented algorithms for path planning in a time-varying environment based on graph methods. The basic task of the introduced path planning algorithms is to find a time-optimal path from a defined start position to a goal position with consideration of the time-varying ocean current for an autonomous underwater vehicle (AUV). Building on this, additional practice-oriented considerations in planning are discussed in this paper. Such points are the discussion of possible methods to accelerate the algorithms and the determination of the optimal departure time. The solutions and algorithms presented in this paper are focused on path planning requirements for the AUV ``SLOCUM Glider''. These algorithms are equally applicable to other AUVs or aerial mobile autonomous systems.
\end{abstract}
\end{frontmatter}

\section{Introduction}\label{sec:Introduction}
This paper is an abridgement of a research fellowship and has been previously published in parts in \cite{Eichhorn2009a,Eichhorn2009b,Eichhorn2010a,Eichhorn2010c}. The following sections review the important results of the study for path planning in a time-varying environment for the autonomous underwater vehicle (AUV) ``SLOCUM Glider''.

Path planning represents an important characteristic of autonomous systems. It reflects the possibility for a planned behaviour during a mission using all current and future information about the area of operation. This planning task will be complicated because of the unknown, inaccurate and varying information. The path planning algorithms presented in this paper are designed considering the mission requirements for the AUV ``SLOCUM Glider''. These gliders have a low cruising speed (0.2 to 0.4 m s$^{-1}$) in a time-varying ocean flow over a long operation range for periods up to 30 days. 

There exists a variety of solutions for the path planning in a time-varying environment, especially for mobile autonomous systems. A generic algorithm was used for an AUV in \cite{Alvarez2004} to find the path with minimum energy cost in a strong, time-varying and space-varying ocean current field. This approach finds a robust solution which will not necessarily correspond with the optimal solution. In \cite{Wang2005}, an adaptive genetic algorithm is presented for determining routes for a large-scale sea area under real-time requirements. The defined fitness function allows the generation of a time-optimal, obstacle-free route with few waypoints. The mixed integer linear programming (MILP) will be used for handling multiple AUVs \cite{Yilmaz2008} or UAVs (Unmanned Aerial Vehicles) \cite{Richards2002}. As the computing time of MILP increases exponentially with the problem size, this approach has limitations in real-sized applications. A solution with a non-linear least squares optimization technique for a path planning of an AUV mission through the Hudson River was presented in \cite{Kruger2007}. The optimization parameters are a series of changeable nodes (\textit{x$_{i}$, y$_{i}$, z$_{i}$, $\Delta{}$t$_{i}$}), which characterize the route. The inclusion of the time intervals \textit{$\Delta{}$t$_{i}$} allows a variation of the vehicle speed during the mission and thus the integration of energy considerations in the optimization. This approach runs the risk of finding only a local minimum. In \cite{Zhang2008} a solution with optimal control to find the optimal trajectory for a glider in a time-varying ocean flow was presented. This approach applied the Nonlinear Trajectory Generation (NTG) algorithm including an ocean current flow B-spline model, a dynamic glider model as well as a defined cost function which is a weighted sum of a temporal and an energy cost. The inclusion of energy requirements using priori known wind information in a graph based path planning for UAVs was discussed in \cite{Kladis2011}. In \cite{Lolla2012}, the level set method for time-optimal path planning in a time-varying flow field is used. In this case, a wave front, starting from the start position is generated. It consists of particles, which describe the most distant position from the vehicle, which can be achieved at a defined time. When the wave front reaches the target position the optimal route will be determined by backtracking the particles. The accuracy of the numerical solution found and the computing time correlate with the defined points to describe the individual wave front lines at certain discrete times. In \cite{Fernandez2010} a modified A*-algorithm was used to find a time optimal path using Regional Ocean Model (ROM) data. This algorithm, called Constant-Time Surfacing A*(CTS-A*), considered the specific glider dynamics under the influence of ocean currents. An A*-algorithm was used in \cite{Pereira2011} to find a minimum risk path for gliders using historical data from the Automated Information System (AIS) for automatically identifying and locating vehicles. 

The chosen pre-defined mesh structure to define the connectivity relations of the several vertices in a geometrical graph has an important influence to find the optimal path in a current field using graph algorithms. This is confirmed in \cite{Garau2009} which is a continuing work of \cite{Garau2005}. Both works use an A* algorithm to find an optimal path in a spatial variability and time-invariant ocean current field. The influence of the mesh structure on the determined path is discussed in Section \ref{subsec:GeometicalGraph}. 
        
The planning algorithms presented in Sections \ref{subsec:ATVEAlgorithm} - \ref{subsec:BothMethods} are based on a modified Dijkstra Algorithm (see \cite{Dijkstra1959}), including the time-variant cost function in the algorithm which will be calculated during the search to determine the travel times (cost values) for the examined edges. This modification allows the determination of a time-optimal path in a time-varying environment. In \cite{Orda1990} this principle was used to find the optimal link combination to send a message via a computer communication network with the shortest transport delay. 

The presented methods to accelerate the path planning algorithms result from trying to determine real mission plans for the AUV ``SLOCUM Glider'' to collect oceanographic data along the Newfoundland and Labrador Shelf \cite{Eichhorn2010a}. The number of edges in the geometrical graph ranges from one hundred thousand to one million for a real mission of duration of 10 days, whereby the sum of the cost function calculations is very time-intensive. This cost function calculations to detect the travel time for an edge are described in Section \ref{sec:CostValue} in detail.

Because the required geometrical graph is not complete as not all vertices are connected by an edge within the graph, the found path has to be smoothed for a real glider mission. This path post processing is a necessary step in real applications (see \cite{Yang2008,Wang2009}). The path smoothing for time varying conditions will be discussed in Section \ref{sec:PathSmooting}.   
      
A fast working algorithm is also a precondition for the detection of an optimal departure time, which is described in Section \ref{sec:OptimalDepartureTime}. A symbolic wavefront expansion (SWE) technique for a UAV in time-varying winds was introduced in \cite{Soulignac2009} to find the time optimal path and additionally the optimal departure time. The path planning algorithms in this paper use a similar principle as is used in the SWE to calculate the time-varying cost function for the several vertices. This includes the arrival time at the several vertices in the cost function calculation during the search. To find the optimal departure time, the SWE and the approach described in this paper use separate solution methods. The reasons are the accurate and fast determination of the optimal departure time, as well as the possible inclusion of uncertainties in the path planning as a result of forecast error variance, accuracy of calculation in the cost functions and a possible use of a different vehicle speed in the real mission than planned \cite{Eichhorn2011a}.

Section \ref{sec:Results} shows the results of the presented algorithms using a simple mathematical model of the Gulf Stream and real netCDF files for a 10-day forecast. Conclusion and future research topics are in Section \ref{sec:Conclusion}.
\section{Graph Algorithm}\label{sec:GraphAlgorithm}
\subsection{Generation of the geometrical graph}\label{subsec:GeometicalGraph}
\begin{figure}[!b]
	\centering
	\vspace{-15pt}
	\includegraphics[width=0.99\columnwidth]{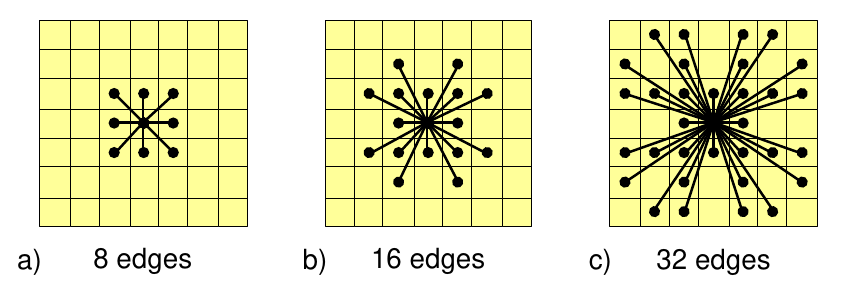}
	\caption{Rectangular grid structure a) 1-sector, b) 2 sector, \newline c) 3 sector}
	\vspace{-15pt}
	\label{img:GridStructure}
\end{figure}

The geometrical graph is a mathematical model for the description of the area of operation with all its characteristics. Therefore defined points (vertices) within the operational area are those passable by the vehicle. In this paper these points define positions in the 2D Euclidean space whereby the geometrical graph is planar. The passable connections between these points are recorded as edges in the graph. Every edge has a rating (cost, weight) which can be the length of the connection, the evolving costs for passing the connection or the time required for traversing the connection. There exist many approaches to describe an obstacle scenario with as few of the vertices and edges as possible, and, to decrease the computing time (visibility and quadtree graph \cite{Eichhorn2004a}). In the case of the inclusion of an ocean current, the mesh structure of the graph will be a determining factor associated with its special change in gradient. In other words, the defined mesh structure should describe the trend of the ocean current flow in the operation area as effectively as possible. A uniform rectangular grid structure is the easiest way to define such a mesh. In the simplest case the edges are the connections between neighbouring obstacle-free sectors; see \autoref{img:GridStructure}(a).
To achieve a shorter and smoother path for mobile robots additional edges to other sectors are implemented in \cite{Ersson2001}; see \autoref{img:GridStructure}(b). The analyses of the found paths in a current field show (see also \cite{Eichhorn2009a,Eichhorn2009b}) that is it important to define a great number of edges with different slopes; see \autoref{img:GridStructure}(c). A further increase of the number of radiated edges leads to increasing lengths which is not practical to describe the change in gradient of the current flow.
\subsection{Graph-based Search-Algorithm}\label{sec:GraphbasedSearchAlgorithm}
The developed search algorithms are all based on the Dijkstra Algorithm \cite{Dijkstra1959} which solves the single-source shortest paths problem on a weighted directed graph. The exact solution by using a Dijkstra algorithm in a time-varying environment requires the inclusion of the time information as an additional dimension in the graph. For instance a 2D geometrical graph acquires additional layers for each defined discrete point of time. That will lead to very large graphs with many vertices and edges as a result of using small time intervals.
To get around this, the algorithms include the time-variant cost function which will be calculated during the search to determine the travel times for the examined edges. The basic algorithm, named TVE (time-varying environment) algorithm, uses only this modification. This algorithm is described in detail in \cite{Eichhorn2009a,Eichhorn2009b}. Its operation is similar to the A*TVE algorithm, which is presented in the next section, using the cost function \textit{d}[\textit{u}] instead of the estimated costs \textit{f}[\textit{u}] in the priority queue \textit{Q}. Methods to accelerate the processing time of the TVE algorithm will be presented in the next sections.
\subsubsection{A*TVE algorithm}\label{subsec:ATVEAlgorithm}
A possible method to accelerate the TVE algorithm is the inclusion of an A* algorithm \cite{Hart1968}.  
The A* algorithm utilizes the Dijkstra algorithm and uses a heuristic function \textit{h}(\textit{u}) to decrease the processing time of the path search. As a heuristic function in the A*TVE algorithm, the travel time $t_{travel}$ from the current vertex \textit{u} at position ${{\mathbf{x}}_{u}}$ to the goal vertex \textit{g} at position ${{\mathbf{x}}_{g}}$ following a straight line based on \cite{Fernandez2010} will be used. Here the travel time will be calculated using the maximum possible speed, as determined by the addition of the vehicle speed through the water ${{v}_{veh\_bf}}$ and the maximum current velocity $v_{current\_max}$ in the operational area over the full mission time:
\begin{equation}
h\left( u \right)={{t}_{travel}}=\frac{\left\| {{\mathbf{x}}_{u}}-{{\mathbf{x}}_{g}} \right\|}{{{v}_{veh\_bf}}+{{v}_{current\_max }}}
\label{eq:hu}
\end{equation}

\autoref{tab:ATVE} shows a comparison between the A* algorithm (left column) and the A*TVE (time-varying environment) algorithm (right column). The syntax of the pseudo-code is adapted from \cite{Siek2002}. The input parameter \textit{G} contains the graph structure with the vertex list and the edge list (\textit{V} and \textit{E}), \textit{s} and \textit{g} are the source and goal vertex and $t_{0}$ includes the starting mission time. The parameter \textit{d} includes the cost list for the several vertices, \textit{f} includes the estimated costs from the source vertex \textit{s} to the goal vertex \textit{g} of the path through the several vertices using the sum of the known cost value \textit{d}[\textit{v}] from the source \textit{s} to the vertex \textit{v} and the value \textit{h(v)} of Eq. \eqref{eq:hu}, and \textit{$\pi{}$} includes the predecessor of each vertex which is used to encode the shortest paths tree \cite{Siek2002}. \textit{Q} is a priority queue that supports the INSERT, EXTRACT-MIN and the DECREASE-KEY operations. The operation EXTRACT-MIN removes the vertex \textit{u} which the least cost value \textit{f}[\textit{u}] in the priority queue \textit{Q}. The operation DECREASE-KEY assigns a new cost value \textit{f} to the vertex \textit{v} in the queue \textit{Q}. The \textit{color} list defines the current state of the vertex in the priority queue \textit{Q}. The allowable states are WHITE, GRAY and BLACK: WHITE indicates that the vertex has not yet been discovered, GRAY indicates that the vertex is in the priority list, and, BLACK indicates that the vertex was checked. The shaded text fields in \autoref{tab:ATVE} highlight the differences between the algorithms.
There are the following three differences:
\begin{enumerate} [leftmargin=*]
	\item The new algorithm does not need the weight list \textit{w} of the edges
to begin the search. The algorithm needs a start time $t_{0}$
when the vehicle begins the mission.
	\item The examination of the edge (\textit{u}, \textit{v}) is only necessary
for \textit{d}[\textit{u}] $<$ \textit{d}[\textit{v}]. It is clear if \textit{d}[\textit{u}]
$\geq{}$\textit{d}[\textit{v}] then \textit{d$_{v}$} $>$ \textit{d}[\textit{v}],
independent of the calculated weight of function \textit{wfunc}. The parameter
\textit{d$_{v}$} includes the calculated cost value for vertex \textit{v} 
by sum of cost value for vertex \textit{u} and the calculated weight of function
\textit{wfunc}.
	\item The algorithm calculates the weight for the edge \textit{w}(\textit{u}, \textit{v}) in a
function \textit{wfunc} during the search. (see Section \ref{subsec:TravelTimeCalculationTimeVarying}, a detailed description
about these calculations will be presented in Section II.b in \cite{Eichhorn2009b}) This function
calculates the travel time to drive along the edge from a start vertex
\textit{u} to an end vertex \textit{v} using a given start time. The
start time to be used will be the current cost value \textit{d}[\textit{u}],
which describes the travel time from the source vertex \textit{s} to
the start vertex \textit{u}.
\end{enumerate}
\begin{table}[!b]
   \centering
   \vspace{-20pt}
   \caption{Pseudo-code of the A* and A*TVE algorithms}
	 \includegraphics[width=0.98\columnwidth]{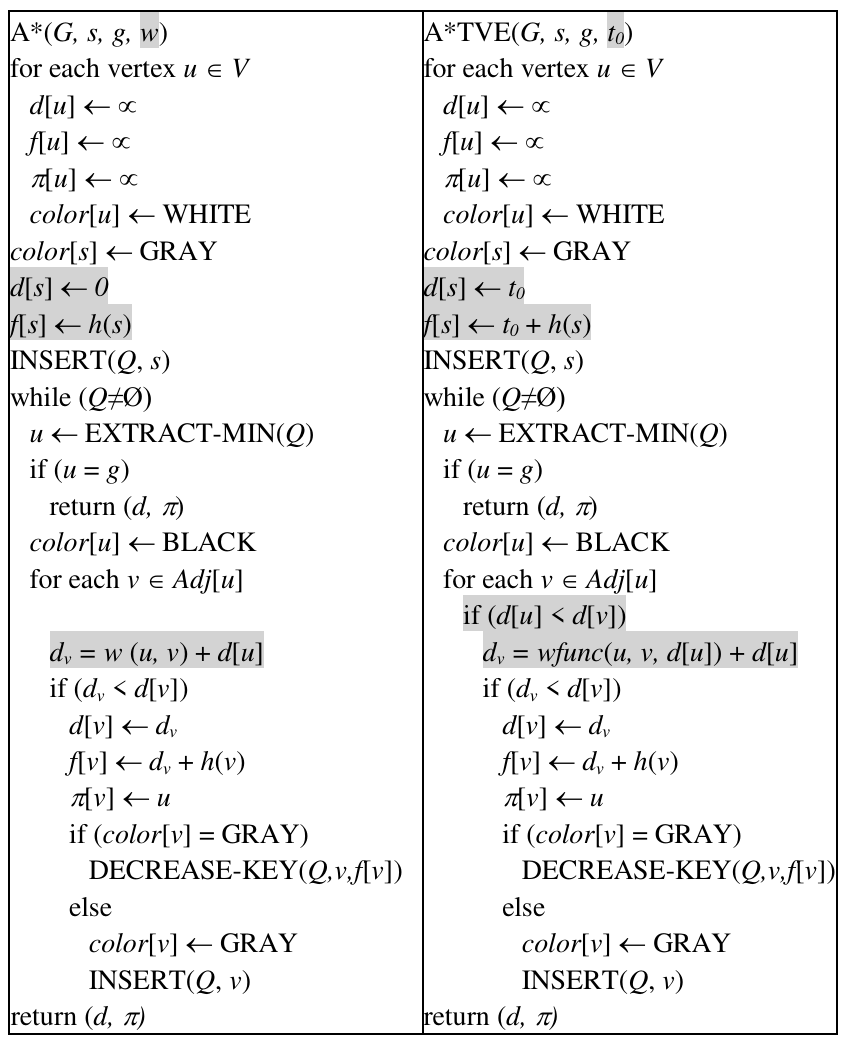}
   \label{tab:ATVE}
\end{table}

\subsubsection{Optimal navigation formula from Zermelo}\label{subsec:Zermelo}
\begin{figure}[b]
	\centering
	\vspace{-15pt}
	\includegraphics[width=0.99\columnwidth]{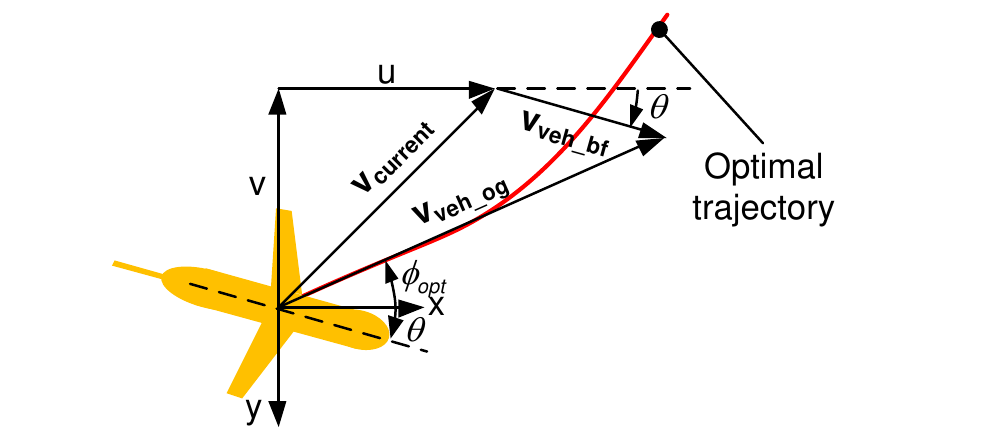}
	\vspace{-22pt}
	\caption{Illustration of the velocities and the angles in glider steering}
	\label{img:VelocitiesZermelo}
\end{figure}

The use of the TVE algorithm to find a time optimal path for the AUV ``SLOCUM Glider'' in time varying ocean flows allows a further possibility to reduce the computing time of the search. This approach uses the optimal navigation formula from Zermelo \cite{Zermelo1931}:
\begin{equation}
\frac{{d\theta }}{{dt}} =  - {u_y}{\rm{co}}{{\rm{s}}^2}\theta  + \left( {{u_x} - {v_y}}
\right){\rm{cos}}\theta {\rm{sin}}\theta  + {v_x}{\rm{si}}{{\rm{n}}^2}\theta 
\label{eq:dtheta}
\end{equation}
with \textit{$\theta{}$} as the heading and \textit{u$_{x}$}, \textit{u$_{y}$}, \textit{v$_{x}$} and \textit{v$_{y}$} as the partial derivatives of the ocean current components \textit{u} and \textit{v}.
The idea to develop this formula came to Zermelo's mind when the airship "Graf Zeppelin" circumnavigated the earth in August 1929 \cite{Ebbinghaus2007}. This formula describes the necessary condition for the control law of the heading \textit{$\theta{}$}, to steer a vehicle in a time-optimal sense through a time-varying current field. The gradient of the resulting optimal trajectory in a fixed world coordinate system is the vehicle velocity over the ground \textbf{\textit{v$_{veh\_og}$}}. This vector is the result of a vector addition of the current vector \textbf{\textit{v$_{current}$}} and the \textbf{\textit{v$_{veh\_bf}$}} vector with vehicle speed through the water \textit{v$_{veh\_bf}$} as norm and heading \textit{$\theta{}$} as direction. The direction of this vector \textbf{\textit{v$_{veh\_og}$}} is the course over the ground (COG) \textit{$\phi{}$}. These relationships are illustrated in \autoref{img:VelocitiesZermelo}. The idea of how to use the optimal navigation formula in the search algorithm as well as the several necessary program steps will be described subsequently.

Assuming that the search algorithm will find the time-optimal path, then the several segments (edges) of this path will match well with the optimal trajectory, which is calculated with optimal control by solving the optimal navigation formula from Zermelo. This assumption means that during the path search only vertices should be considered where the connections (edges) comply with the optimal navigation formula. This compliance is required where the transition that is the change of direction between two adjacent edges is matched with Eq. \eqref{eq:dtheta}.

\begin{figure}[t]
	\centering
	\includegraphics[width=0.99\columnwidth]{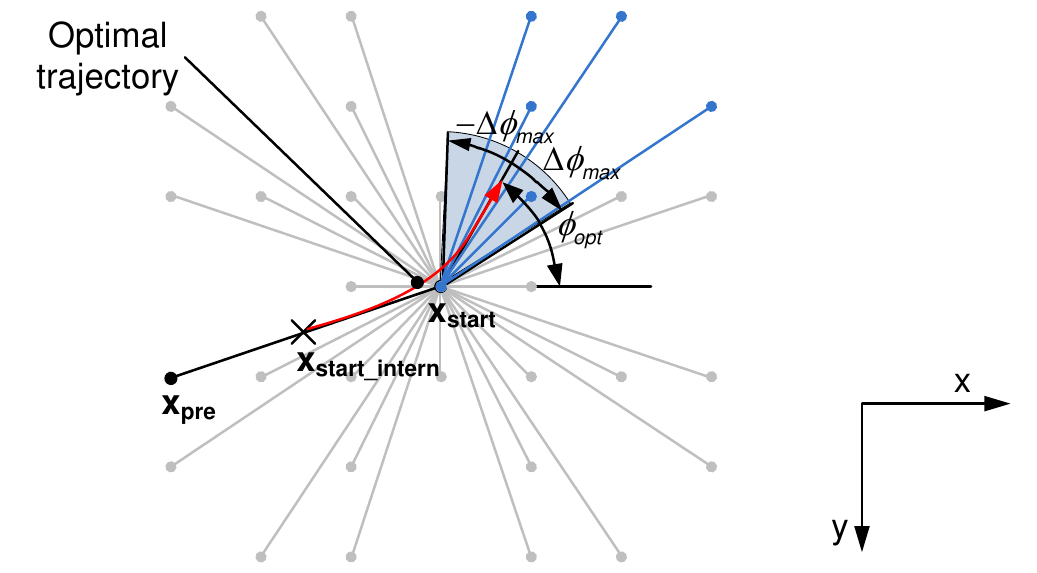}
	\vspace{-15pt}
	\caption{Resultant angle range to define the examined successor vertices using optimal navigation formula from Zermelo}
	\vspace{-15pt}
	\label{img:AngleRange}
\end{figure}

The determination of the optimal path direction \textit{$\phi{}$$_{opt}$} on position \textit{x$_{start}$} required a simulation of the optimal trajectory by starting on the middle position of the previous edge by \textit{x$_{start\_intern}$} (see section II.D in \cite{Eichhorn2010c}). The calculated path direction \textit{$\phi{}$$_{opt}$}  will be used to select possible successor edges with the end vertex \textit{v} under consideration of an angle range $\pm{}$$\Delta{}$\textit{$\phi{}$$_{max}$} (see \autoref{img:AngleRange}). This range considers the maximal possible angle between two adjoined edges and the fact that the path direction \textit{$\phi{}$$_{opt}$} is only an average value along the path and is predetermined through the given numbers of possible edges from the differences in slopes according to the chosen mesh structure (see \autoref{img:GridStructure}).

This approach incorporates a pre-selection of promising successor vertices with the goal to decrease the number of cost function calls \textit{wfunc} during the search. \autoref{img:AngleRange} shows the principle idea of the approach using a 3-sector rectangular grid structure which is described in Section \ref{subsec:GeometicalGraph}. By using such a structure, 31 successor vertices are possible from which the approach selects only five. This occurs in the best case (for all examined vertices \textit{v} is \textit{d}[\textit{u}] $<$\textit{ d}[\textit{v}]) with a resulting reduction of the called \textit{wfunc} to 83 \% using this ZTVE algorithm.
\subsubsection{The use of both methods}\label{subsec:BothMethods}
The use of both methods together, the A* algorithm (Section \ref{subsec:ATVEAlgorithm}) and the optimal navigation formula from Zermelo (Section \ref{subsec:Zermelo}) in the TVE algorithm combines the two acceleration mechanisms and produces a larger reduction in the computing time than with either method alone. \autoref{tab:ZATVE} shows this algorithm, named ZA*TVE, with a few explanations. The modifications to the TVE algorithm are highlighted. The letters which appear in the explanation column refer to the used method (A*: A* algorithm; Z: Zermelo's formula). According to Eq. \eqref{eq:dtheta}, the function CAL-OPTDIR calculates the optimal path direction \textit{$\phi{}$$_{opt}$} on position \textit{u} to the time \textit{d}[\textit{u}] using the direction of the edge with the predecessor vertex \textit{$\pi{}$}[\textit{u}] as start vertex and the current examined vertex \textit{u} as the end vertex which should reflect the average optimal COG.

At this point additional acceleration possibilities should be discussed briefly. The first possibility includes the selective reduction of the search area to decrease the number of examined vertices during the search. To do this a first search run uses a graph with a large grid size and/or a simple grid structure (see Section \ref{subsec:GeometicalGraph}). Around the found path a new geometrical graph will be generated, similar to a pipe. This graph will have a fine grid size and/or a complex grid structure and will be used in a second run to find the optimal path. A modification of the upper approach is the use of a simple cost function in the first search run and the use of an accurate glider-model in the cost function for the second run.

\begin{table}[!ht]
   \centering
   \caption{Pseudo-code of the ZA*TVE algorithm}
	 \includegraphics[width=0.98\columnwidth]{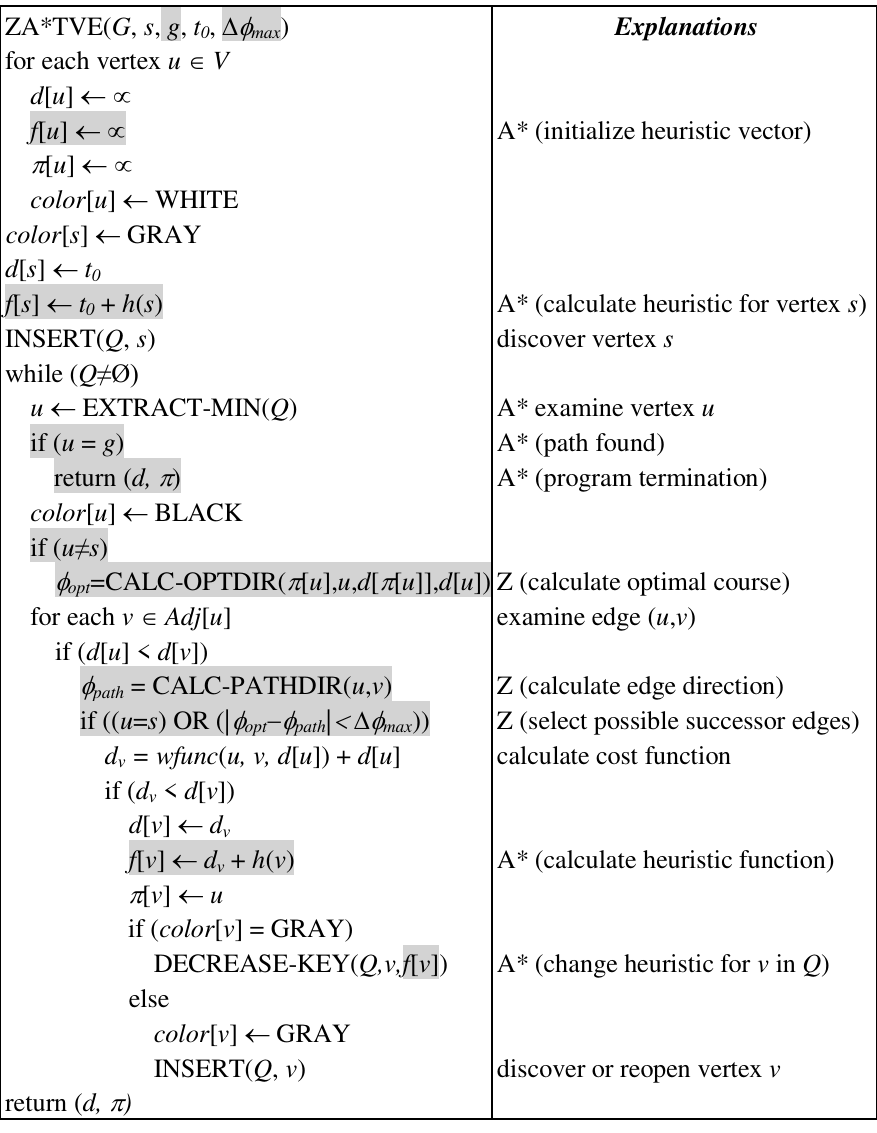}
   \label{tab:ZATVE}
\end{table}

\section{Calculation of the cost value}\label{sec:CostValue}
This section describes the necessary equations and algorithms to determine the travel time ${{{t}^{i}}_{path}}$ for the several path segments using information about the ocean current.
\subsection{Travel time calculation}\label{subsec:TravelTimeCalculation}
\begin{figure}[t]
	\centering
	\includegraphics[width=0.99\columnwidth]{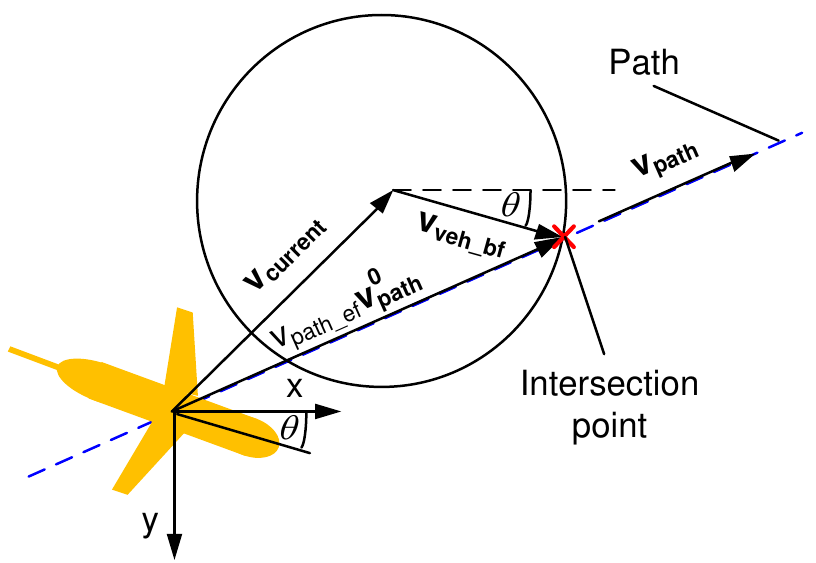}
	\caption{Definition of the velocities}
	\label{img:DefinitionOfVelocities}
	\vspace{-10pt}
\end{figure}

The travel time can be calculated for the \textit{i$^{th}$} edge by formation of the quotient of the distance $s_{path}$ and the speed ${{v}_{path\_ef}}$, with which the vehicle travels on the path in relation to a fixed world coordinate system:
\begin{equation}
t_{path}^{i}=\frac{s_{path}^{i}}{v_{path\_ef}^{i}}
\label{eq:tpath}
\end{equation}
This speed ${{v}_{path\_ef}}$ depends on the vehicle speed through the water ${{v}_{veh\_bf}}$ (cruising speed), the magnitude and the direction of the ocean current vector as well as the direction of the path $\mathbf{v}_{path}^{0}$. This speed can be determined by the intersection point between a line and a circle (2D) and/or sphere (3D) \cite{Schneider2003} based on \autoref{img:DefinitionOfVelocities} according to the following relation:
\begin{equation}
  \begin{aligned}
   &line\text{: } \mathbf{x}\left( {{v}_{path\_ef}} \right)={{v}_{path\_ef}}\mathbf{v}_{path}^{0}\\ 
   &circle/spheres\text{: } v_{veh\_bf}^{2}={{\left\| \mathbf{x}-{{\mathbf{v}}_{current}} \right\|}^{2}} \\ 
\end{aligned}
\label{eq:intersection}
\end{equation}

\begin{equation}
\begin{aligned}
  &disc={{\left( \mathbf{v}{{_{path}^{0}}^{\mathbf{T}}}\cdot {{\mathbf{v}}_{current}} \right)}^{2}}+v_{veh\_bf}^{2}-{{\mathbf{v}}_{current}}^{\mathbf{T}}\cdot {{\mathbf{v}}_{current}} \\ 
\end{aligned}
\label{eq:disc}
\end{equation}

\begin{figure}[b]
	\centering
	\includegraphics[width=0.99\columnwidth]{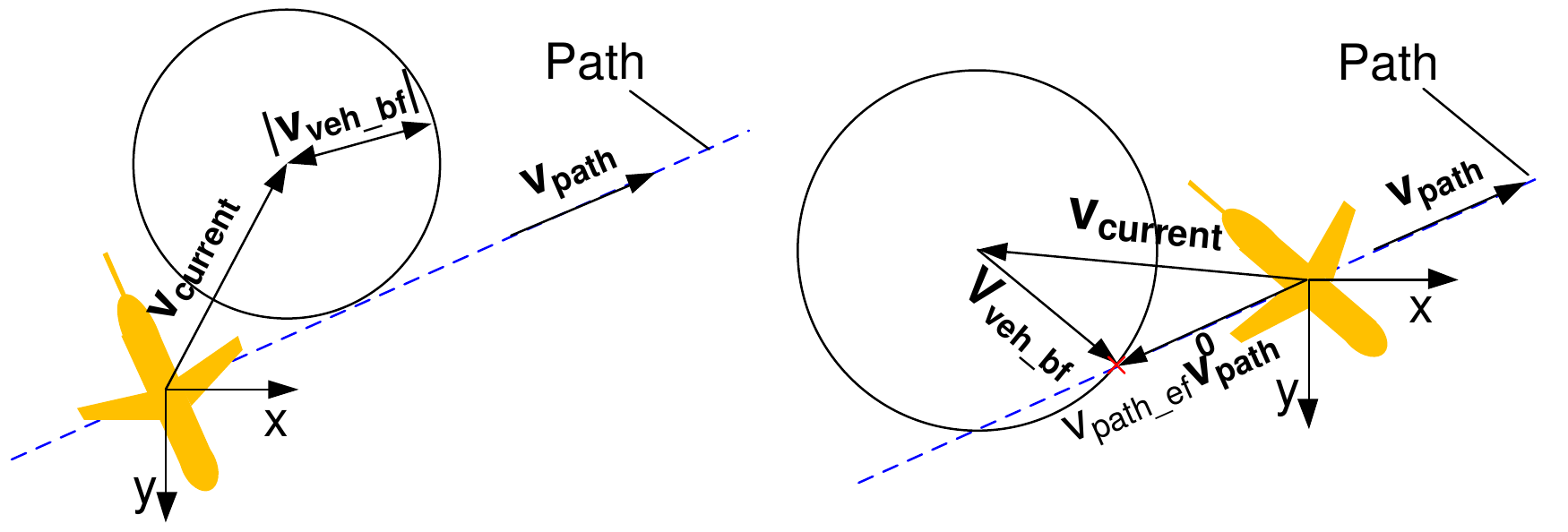}
	\caption{a) Negative discriminant b) ${{v}_{path\_ef} < 0}$}
	\label{img:NegativeDiscriminant}
\end{figure}

If the discriminant ${disc}$ in equation Eq. \eqref{eq:disc} becomes negative, ${{v}_{path\_ef}}$ has not a real solution, so that the value will be defined as ${NaN}$ (\textbf{N}ot \textbf{a} \textbf{N}umber):
\begin{equation}
\begin{aligned}
 & {{v}_{path\_ef}}=\left\{ 
 	\begin{aligned}
 	& \mathbf{v}{{_{path}^{0}}^{\mathbf{T}}}\cdot {{\mathbf{v}}_{current}}+\sqrt{disc}&&, \text{ for }disc>0 \\ 
 	& NaN&&, \text{ otherwise.} \\ 
	\end{aligned} \right. \\ 
\end{aligned}
\label{eq:v_path_ef}
\end{equation}
 It means that the vehicle can no longer be held in that path, the path is not feasible; see \autoref{img:NegativeDiscriminant} (a). If the speed ${{v}_{path\_ef}}$ is negative the vehicle is still on the path, but moving backwards; \autoref{img:NegativeDiscriminant} (b). Both cases must be considered by setting a large numerical value for the edge weight. Thus, such paths are excluded in the search and it does not come to a situation that the vehicle encounters a strong backwards current and leaves the path.
\subsection{Travel time calculation in time-varying ocean flow}\label{subsec:TravelTimeCalculationTimeVarying}
The determination of the travel time according to equation Eq. \eqref{eq:tpath} works only if the ocean current is constant along the path, or through an appropriate choice of the mesh sizes of the graph for a location-varying ocean current. In the case of a time-varying ocean current or a too coarse mesh structure used in conjunction with a location-varying ocean current, the speed ${{v}_{path\_ef}}$ will be changed depending on the current \textbf{v\textit{$_{current}$}} along the path element.
The used algorithm to solve this problem is based on a step size control for efficient calculation of numerical solutions of differential equations \cite {Hundsdorfer2003}. The step size \textit{h} is here not the time as used in numerical solvers but is a segment of the path element. So the path element will be shared within many segments, for which Eq. \eqref{eq:tpath} using Eq. \eqref{eq:v_path_ef} can be solved. The current \textbf{v\textit{$_{current}$}} in Eq. \eqref{eq:v_path_ef} is the arithmetic mean of the two velocities at the begin and the end of the several shared element. Detailed information about the algorithm is described in Section III.B \cite{Eichhorn2009b}. 
\subsection{Glider dive profile cost function}\label{subsec:GliderDiveProfile}
\begin{figure}[b]
	\centering
	\includegraphics[width=0.99\columnwidth]{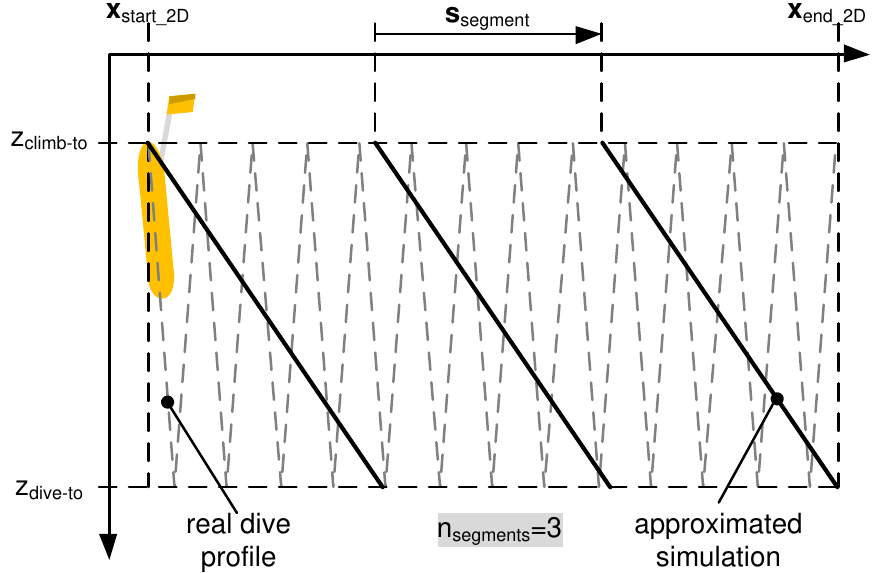}
	\caption{Simplified dive profile along a path element}
	\label{img:DiveProfile}
\end{figure}

The glider dive profile is specified by its locomotion principle. By changing its buoyancy, the glider is able to descend (dive) and ascend (climb). The result is a saw-tooth vertical profile as shown schematically in \autoref{img:DiveProfile}. The exact simulation of such a dive profile is computationally time-intensive and so is impractical because of the number of edges in the geometrical graph, which range from one hundred thousand to one million. Conversely, the knowledge of the glider's behaviour in every passable depth is necessary for the
planning and makes it possible that the mission planning can avoid regions with an adverse surface or seabed current.

To include the depth-varying ocean current information in the cost function (presented in Section \ref{subsec:TravelTimeCalculation}) the path element is divided into several path segments. The number of the segments \textit{n$_{segments}$} is defined by the step size \textit{h}. This number shall consider the changeable ocean current conditions along the path at every passable depth. In each segment the glider dives from the \textquotedblleft{}climbto\textquotedblright{} depth \textit{z$_{climb-to}$} until the
\textquotedblleft{}diveup\textquotedblright{} depth \textit{z$_{dive-up}$}; see \autoref{img:DiveProfile}. The calculated travel time \textit{t$_{travel}$} for the segment will correspond approximately to the travel time which the glider needs to travel along every segment of the saw-tooth profile. \autoref{img:DiveProfile} shows the simplified dive profile in
comparison to the real saw-tooth profiles. The details of the algorithm to calculate the travel time are included in \cite{Eichhorn2010a}.
\subsection{Ocean current determination}\label{subsec:OceanCurrentDetermination}
The cost function to calculate the travel time operates like a numerical simulation (see Section \ref{subsec:TravelTimeCalculation} and \ref{subsec:TravelTimeCalculationTimeVarying}). This simulates the vehicle driving along the path from a defined start position \textbf{x}\textit{$_{start}$} at start time \textit{t$_{start}$} to the end position \textbf{x}\textit{$_{end}$} under consideration of the local ocean
current. During such a simulation the cost function requires a large amount of local ocean current information, which is generated in the ocean current model. Thereby the Cartesian position \textbf{x}, the depth \textit{z} and the time \textit{t} are passed from the cost function to the ocean current model. The model returns a two dimensional ocean current vector \textbf{v$_{c}$}~= ~[\textit{u~v}]. To determine the travel time, accurate ocean current information along the path element is required. This ocean current information will be provided in this research project through the DFO's (Department of Fisheries and Oceans) Canada-Newfoundland Operational Ocean Forecast System (C-NOOFS) in the form of netCDF files, generated in a numerical model. At present it is not possible to directly couple the search algorithm and the numerical model because of time and computational constraints.
\subsubsection{Preparation of the netCDF-Files}\label{PreparationNetCDFFiles}
The Network Common Data Format (NetCDF) is a binary data format for array-oriented scientific data \cite{Unidata2011} which is commonly used for climatology, meteorology and oceanography applications. The C-NOOFS
provides the ocean current data at various depths (0 to 5700 m) for the entire Northwest Atlantic with a resolution of approx. 6 km in the region of interest, every 6 h for a 10-day forecast in geographical coordinates. To use these data in the ocean current model, they need to be extracted out of the region of interest in a Cartesian coordinate system as reference. These modifications can be addressed by using the FIMEX (File Interpolation, Manipulation and EXtraction) library \cite{Fimex2011}.
\subsubsection{Multi-dimensional interpolation scheme}\label{MultiDimensional}
\begin{figure}[t]
	\centering
	\includegraphics[width=0.99\columnwidth]{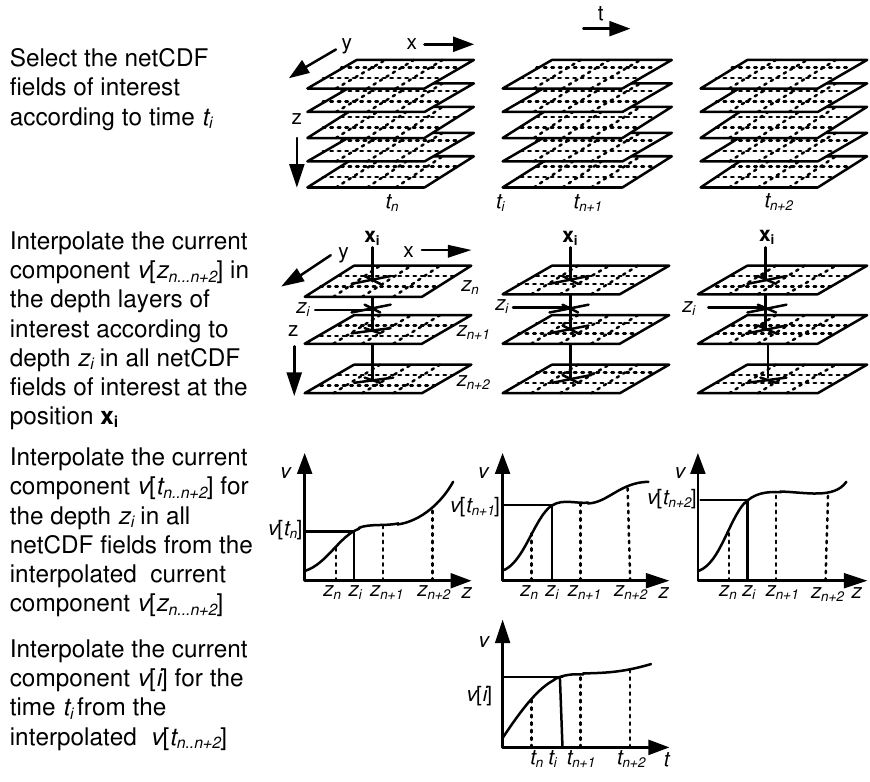}
	\caption{Steps to interpolate an ocean current component
\textit{v}[\textit{i}] at position \textbf{x}[\textit{i}], depth
\textit{z$_{i}$} and time \textit{t$_{i}$} from the netCDF fields}
	\label{img:InterpolationSteps}
	\vspace{-10pt}
\end{figure}

Since the ocean current data coming from the forecasting system as data files will be provided only at discrete times and positions with a coarser time and length scale than is required to generate an efficient path, a multi-dimensional interpolation scheme will be utilized to generate the desired data. \autoref{img:InterpolationSteps} shows the scheme for the ocean current component \textit{v} in overview. The first interpolation step uses a two-dimensional interpolation function from the FIMEX library to extract the ocean current information for the several depth layers. A Nearest-Neighbour, a Bilinear and a Bicubic interpolation method are available. The interpolations via the depth and time dimensions occur separately using one-dimensional interpolation functions. Nearest-Neighbour, Linear, Cubic and Akima interpolation are possible. The first two methods require two fields (for time \textit{t}) or layers (for depth \textit{z}), the other methods require minimal three, optimal five fields or layers in order to generate the ocean current component \textit{v }at the defined position \textbf{x$_{i}$}, at the depth \textit{z$_{i}$} and at the time \textit{t$_{i}$}. The implementation of the Akima interpolation \cite{Akima1970} can make allowance for an abrupt change of ocean current conditions in case of tides or of different depth streams. 

\section{Path smoothing}\label{sec:PathSmooting}
\begin{figure}[!ht]
	\centering
	\includegraphics[width=0.99\columnwidth]{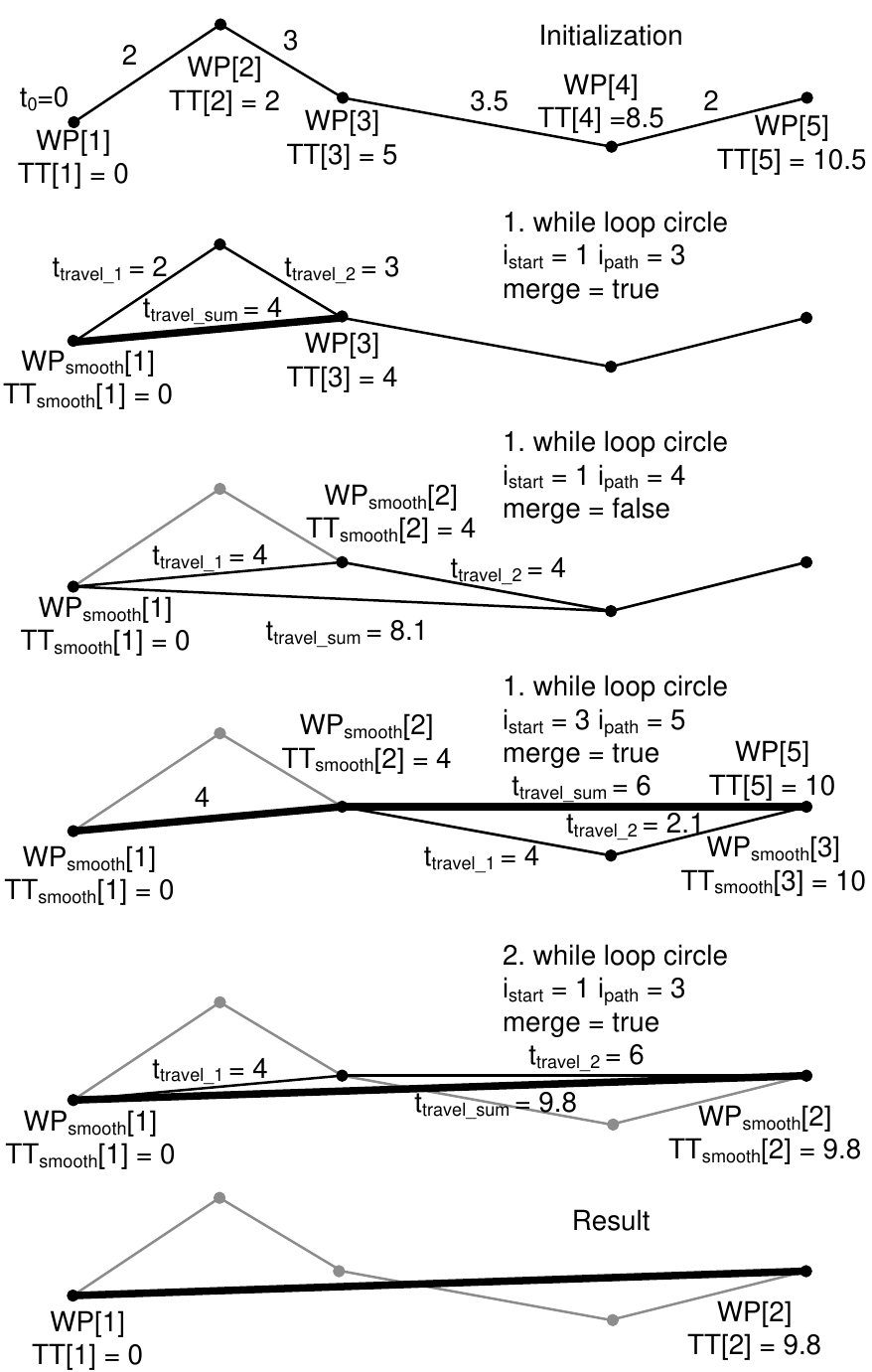}
	\vspace{-20pt}
	\caption{Several steps to smooth a path}
	\label{img:StepsToSmooth}
	\vspace{-15pt}
\end{figure}

The required geometrical graph for the search algorithm is not complete as not all vertices are connected by an edge within the graph (see Section \ref{subsec:GeometicalGraph}). A complete directed graph on \textit{n} vertices has \textit{n}(\textit{n}-1) edges, which would evolve into a very large graph and a long computing time for the path search. So, the search algorithm can consider in its search only the edges which are included in the non-complete graph. The paths found are characterized by many several path segments with change of directions. The run of such a path is stair- or wiggle-shaped, which a glider cannot follow. Therefore a method to smooth the candidate path will be presented as follows.
\autoref {tab:PathSmoothing} includes the details of the algorithm to smooth the path
under consideration in the time-varying environment.   
The candidate path is described by a list of waypoints \textit{WP}. The algorithm verifies the start point \textit{WP}[\textit{i$_{start}$}] of the list with the subsequent waypoints \textit{WP}[\textit{i$_{path}$}] of a direct connection (\textbf{III}), with the goal of a quicker arrival at this point by using the several path elements (\textbf{IV}). Verification of the arrival time \textit{TT}[\textit{end}] of the goal point \textit{WP}[\textit{end}] from the tested waypoint \textit{WP}[\textit{i$_{path}$}] using the existing subsequent waypoints (\textbf{V}) also occurs. This second verification through the time-varying environment is necessary and ensures that the merge of path elements indeed leads to a quicker arrival at a local waypoint, but leads to a later arrival time at the goal point. This is possible even though the ocean current situation is changing dynamically. The merging begins by the third waypoint (\textbf{II}) and will be executed until one of the two verifications is satisfied or the goal point is reached. In the case of a positive verification (\textit{merge} = \textit{false}), the present waypoint \textit{WP}[\textit{i$_{path}$}]{\footnotesize} will be stored in the new waypoint list and a new merge will begin at the precedent waypoint (\textbf{VI}). The result is a waypoint list \textit{WP$_{smooth}$} with fewer waypoints in the verified waypoint list \textit{WP}.   If obstacles are encountered, the function TRAVELTIME also calculates the intersections of the obstacles with the several path elements. In case of a collision situation, the resulting time value has a large numerical value. 
\begin{table}[!b]
   \centering
   \vspace{-10pt}
   \caption{Algorithm for path smoothing in time-varying environment}
	 \includegraphics[width=0.99\columnwidth]{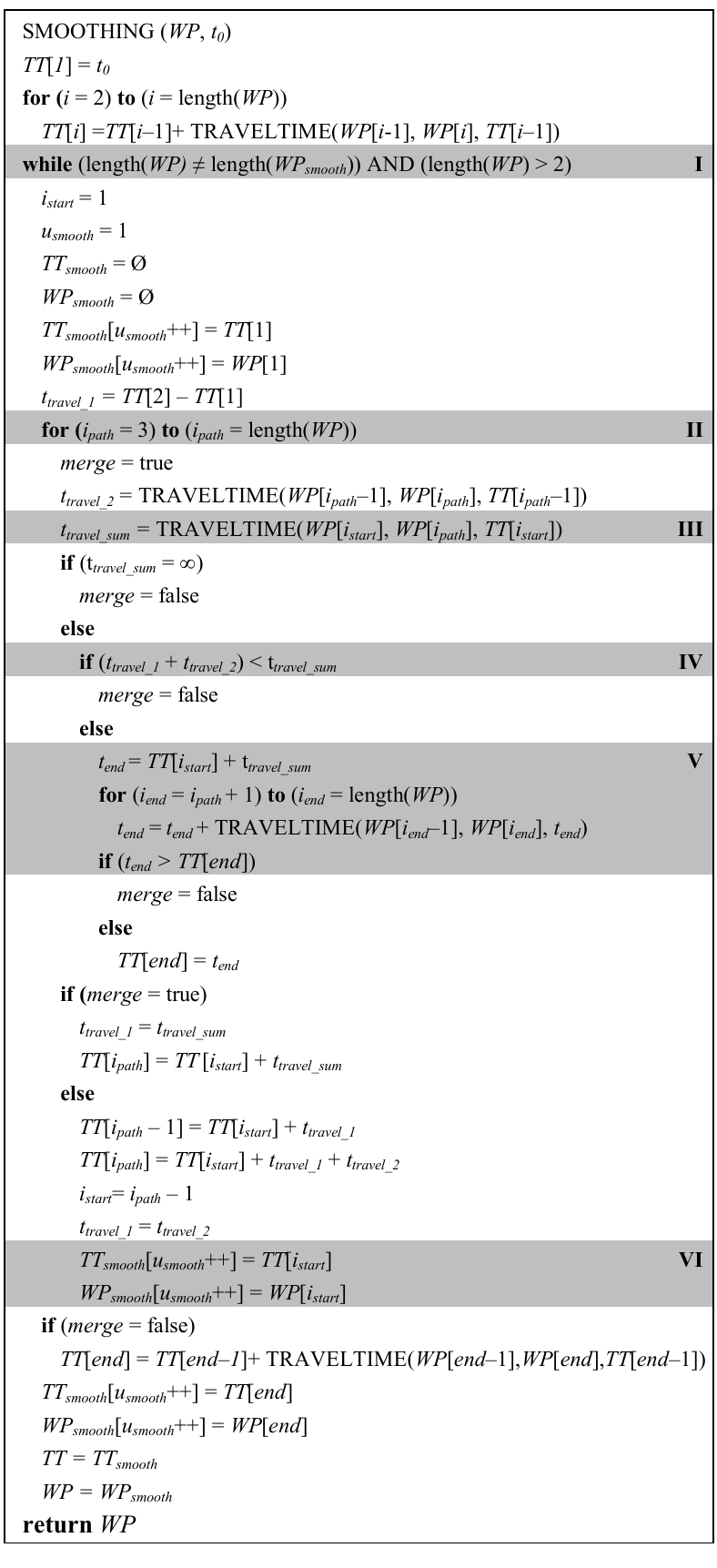}
	 \vspace{-20pt}
   \label{tab:PathSmoothing}
\end{table}
The above described procedure will be repeated until the number of waypoints is constant between two sequent loops (\textbf{I}).
\autoref{img:StepsToSmooth} shows an example for a better understanding.
\section{Detection of the optimal departure time}\label{sec:OptimalDepartureTime}
A practice-relevant requirement for optimal path planning for the AUV ``SLOCUM Glider'' is the determination of the optimal departure time. So is it very difficult to start a glider mission near the coast in the presence of strong tides. Through the low cruising speed (0.2 to 0.4 m s$^{-1}$) and a false chosen start time in combination with a strong flowing tide, it is possible that the glider will make poor forward progress or drift back to the shore. Another scenario is a bad weather situation or a temporary adverse ocean current condition in the region of interest.
\subsection{Idea}\label{sub:Idea}
The function to describe the relationship between the travel time \textit{t$_{trav}$} and departure time \textit{t$_{dep}$} consists of an independent single pair of variants. This means that to determine
the travel time for a certain departure time, the knowledge of travel times with a lesser departure time is not necessary. Because of this, it is possible to reproduce the principle run of the curve \textit{t$_{trav}$}~=~f(\textit{t$_{dep}$}) using a smaller number of defined departure times \textit{t$_{dep\_i}$}, distributed in the time window of interest, to find the corresponding travel times
\textit{t$_{trav\_i}$}. In an additional step the region of the global minimum can be localized, to detect the optimal departure time using a root-finding algorithm. The algorithmic details will be described in the next section.
\subsection{Algorithm}\label{sub:Algorithm}
\begin{figure}[!hb]
	\centering
	\includegraphics[width=0.99\columnwidth]{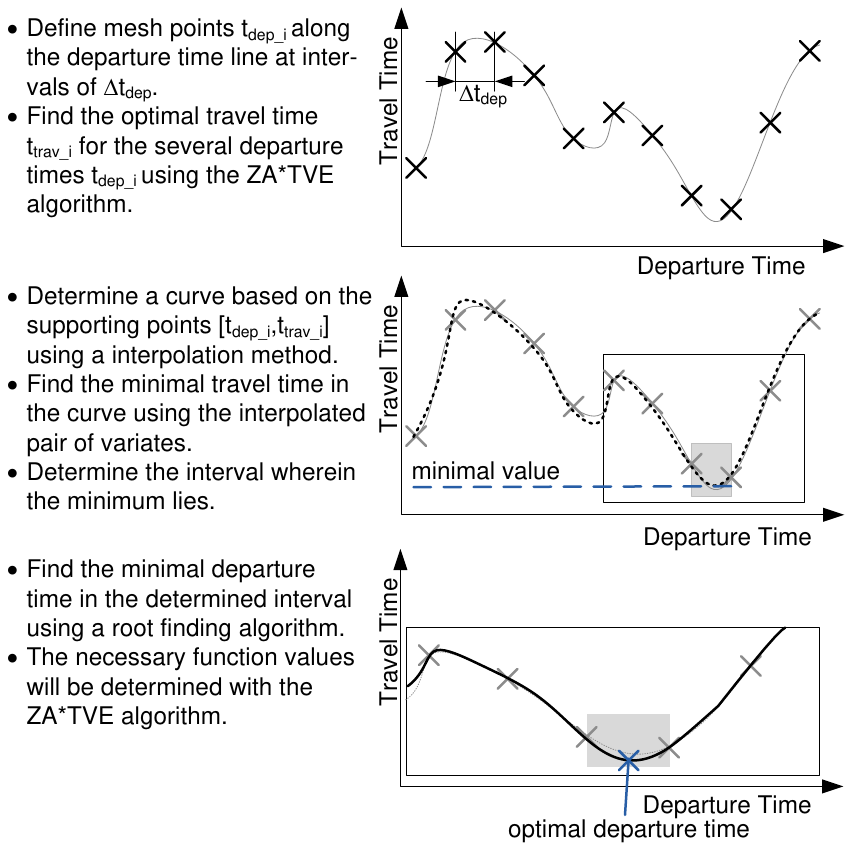}
	\caption{Steps to find the optimal departure time}
	\label{img:OptimalDepartureTime}
	\vspace{-15pt}
\end{figure}

The detection of the optimal departure time occurs in three steps. \autoref{img:OptimalDepartureTime} displays an overview of the scheme to determine the optimal departure time. The first step creates supporting points for the curve \textit{t$_{trav}$}~= f(\textit{t$_{dep}$}) at intervals of \textit{$\Delta{}$t$_{dep}$}. The choice of the interval width is based on the run of the curve and should reflect the positions of the local minima.

These supporting points will be provided in a second step to create the approximated run of the curve using an interpolation method. The studies in this research field favour the Akima interpolation \cite{Akima1970}.
This method provides the best fitting to the real curve and tries to avoid overshoots, which would indicate a nonexistent minimum. The determination of the interval wherein the global minimum of the approximated curve lies is the precondition for the last step. 

Here a one-dimensional root-finding algorithm will be used to find the
optimal departure time. Thereby a path search using the ZA*TVE algorithm will be running alongside every function call to find the travel time for the given departure time. For root-finding algorithms, root-bracketing algorithms will be used. These algorithms work without derivatives and find the root through iterative decreasing of the interval until a desired tolerance is achieved, wherein the root lies. Golden section search \cite{Kiefer1953}, Fibonacci search \cite{Ferguson1960} and Brent's algorithm \cite{Brent1973} were tested. Brent's algorithm has the best performance and will be favoured.

\subsection{Possible Modifications and critical notes}\label{sub:PossibleModifications}
The above described algorithm calls the search algorithm multiple times, which correlates directly with the processing time. A possibility to reduce the processing time will be discussed briefly. Because the localized global minimum in the second step represents only the rough position, the supporting points used in the interpolation do not have to be accurate. This means that in order to detect these points a graph with a larger grid size and/or a simpler grid structure can be used. The result is a generated graph with fewer edges to be examined during the search, which leads to a more rapid calculation of the approximated travel time for the given departure time.

The multiple calls of the search algorithm can be calculated independently at the same time on separate processor cores of a multi-core computer. The analyses of the possibilities for parallelization and the programmable implementation are current work fields \cite{Eichhorn2011a}.

Use of this approach in a real application requires recognition of the fact that only a limited extent of the forecast window will be available. So, the possible mission window is narrowed down to the period between the considered departure time and the forecast horizon. In the application presented in \cite{Eichhorn2010a} the forecast window for the ocean currents is 10 days. This means that if one starts a mission on the ninth day only a one day mission can be planned. Another aspect is the delayed supply of the data of interest in the case of a later start time.
\section{Results}\label{sec:Results}
\subsection{The selected test function for a Time-Varying Ocean Flow}\label{sub:TestFunction}
The function used to represent a time-varying ocean flow describes a meandering jet in the eastward direction, which is a simple mathematical model of the Gulf Stream \cite{Cencini1999,Alvarez2004}. This function was applied in \cite{Eichhorn2009b,Eichhorn2010a} and \cite{Eichhorn2010c} to test the TVE algorithm and its modifications and in \cite{Eichhorn2011a} to show the influence of the methods to realize fast search algorithms and to find suboptimal paths using uncertain information. The stream function is
\begin{equation}
\phi (x,y) = 1 - {\rm{tanh}}\left( {\frac{{y -
B(t)\cos \left( {k\left( {x - ct} \right)} \right)}}{{{{\left( {1 +
{k^2}B{{(t)}^2}{{\sin }^2}\left( {k\left( {x - ct} \right)} \right)}
\right)}^{\frac{1}{2}}}}}} \right)
\label{phixy}
\end{equation}
which uses a dimensionless function of a time-dependent oscillation of
the meander amplitude
\begin{equation}
B(t)={{B}_{0}}+\varepsilon \cos (\omega t+\theta )\text{ }
\label{Bt}
\end{equation}
and the parameter set \textit{B$_{0 }$}= 1.2, \textit{$\epsilon{}$}  =
0.3, \textit{$\omega{}$} = 0.4, \textit{$\theta{}$}  = $\pi{}$/2, 
\textit{k} = 0.84 and \textit{c }= 0.12 to describe the velocity field:
\begin{equation}
u(x,y,t) =  - \frac{{\partial \phi }}{{\partial y}}{\rm{   }}\text{ }  
v(x,y,t) = \frac{{\partial \phi }}{{\partial x}}
\label{uxyt}
\end{equation}
The dimensionless value for the body-fixed vehicle velocity v\textit{$_{veh\_bf}$} is 0.5. This test function makes it possible to show very transparently how a path planning algorithm works with uncertain information. The exact time optimal solution was found by solving a boundary value problem (BVP) with a collocation method bvp6c \cite{Hale2008} in MATLAB. The three
ordinary differential equations (ODEs) include the two equations of motion:
\begin{equation}
	\begin{array}{l}
		\frac{{dx}}{{dt}} = u + {v_{veh\_bf}}\cos \theta \\
		\frac{{dy}}{{dt}} = v + {v_{veh\_bf}}\sin \theta
	\end{array}
	\label{dxdz}
\end{equation}
and the optimal navigation formula from Zermelo in Eq. \eqref{eq:dtheta}.
\begin{figure}[!b]
	\centering
	\vspace{-20pt}
	\includegraphics[width=0.95\columnwidth]{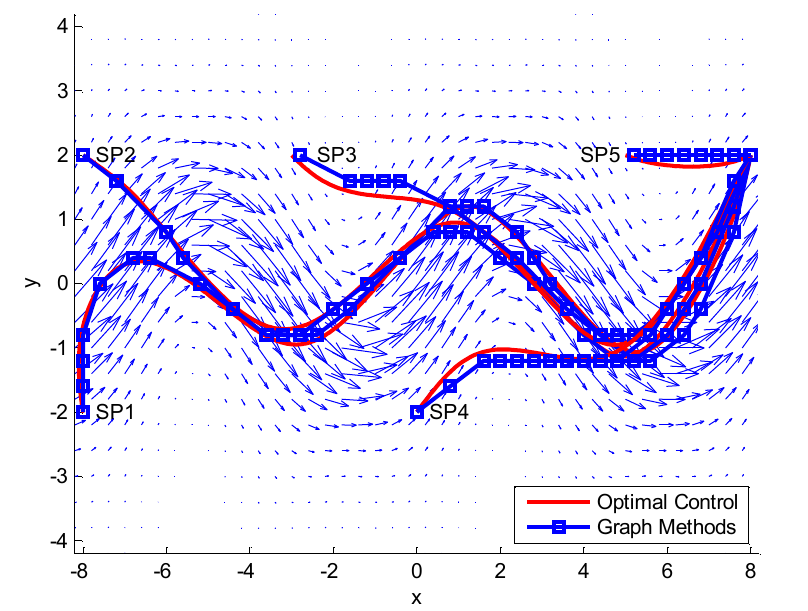}
	\vspace{-10pt}
	\caption{Time optimal paths through a time-varying ocean field using Optimal Control and the Graph 
	         Methods for different start positions}
	\label{img:OptimalPaths}
\end{figure}
\subsection{Comparison between the methods to accelerate the TVE algorithm}\label{sub:Comparison}
This section presents the results of the methods to accelerate the TVE algorithm which are described in Section {\ref{subsec:ATVEAlgorithm} - \ref{subsec:BothMethods} using the time-varying ocean flow test function of the previous section. For the test cases, five different start positions were distributed in the whole area of operation as shown in \autoref{img:OptimalPaths}. All the graph-based methods use the same
graph and hence produce identical paths. \autoref{img:OptimalPaths} shows the five paths found using optimal control and the graph methods. For the graph methods, the rectangular 3-sector grid structure with a grid size of 0.4 was used (see \autoref{img:GridStructure}(c)). \autoref{img:CostFunctionCalls} shows the necessary number of cost function calls (CFC) using the several methods for the five start positions. All these results are included in \autoref{tab:DifferentSearchMethods}. The examination of the current model calls should reflect their ratio to the cost function calls, which is important in the case of computing intensive ocean current calculations. Using the A*TVE algorithm (see Section \ref{subsec:ATVEAlgorithm}, the number of function calls correlates directly with the distance between the start and the goal position.
\begin{table}[!htp]
  \centering
  \caption{Results of the different search methods}
	\includegraphics[width=0.95\columnwidth]{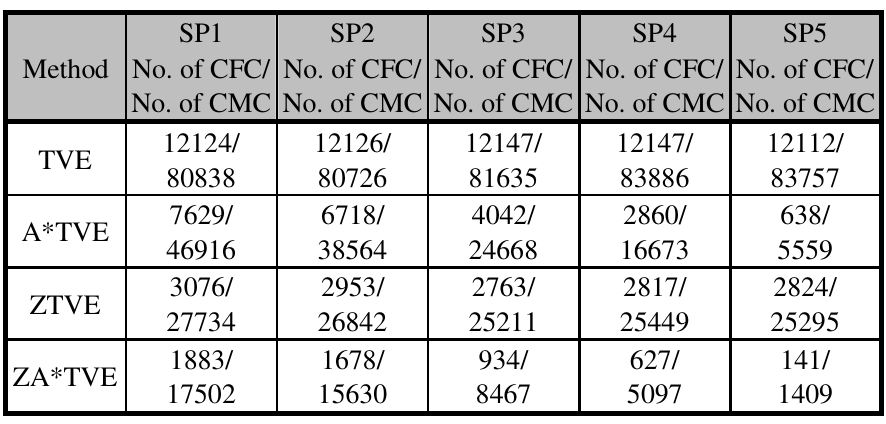}
  \label{tab:DifferentSearchMethods}
  \vspace{-4pt}
\end{table}
\begin{figure}[h]
	\centering
	\includegraphics[width=0.99\columnwidth]{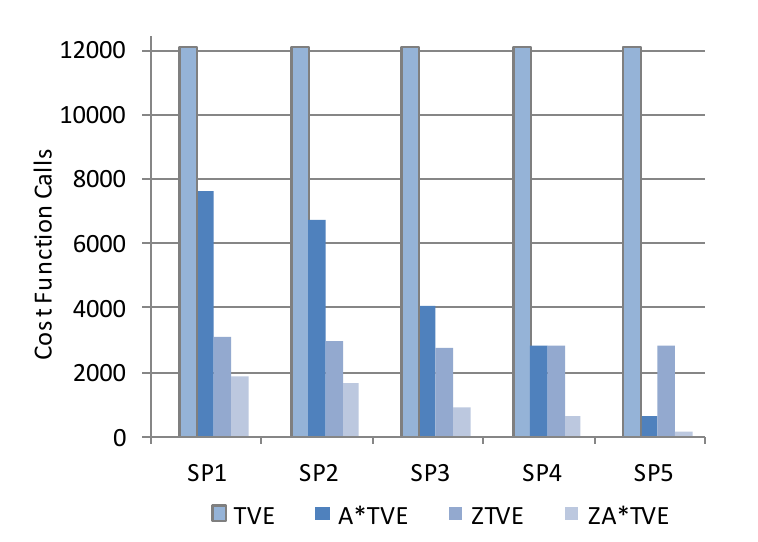}
	\caption{Cost function calls for the various methods with different start positions}
	\vspace{-5pt}
	\label{img:CostFunctionCalls}
	\vspace{-20pt}
\end{figure}
\begin{figure}[!hb]
\begin{center}
	 \includegraphics[width=0.99\textwidth]{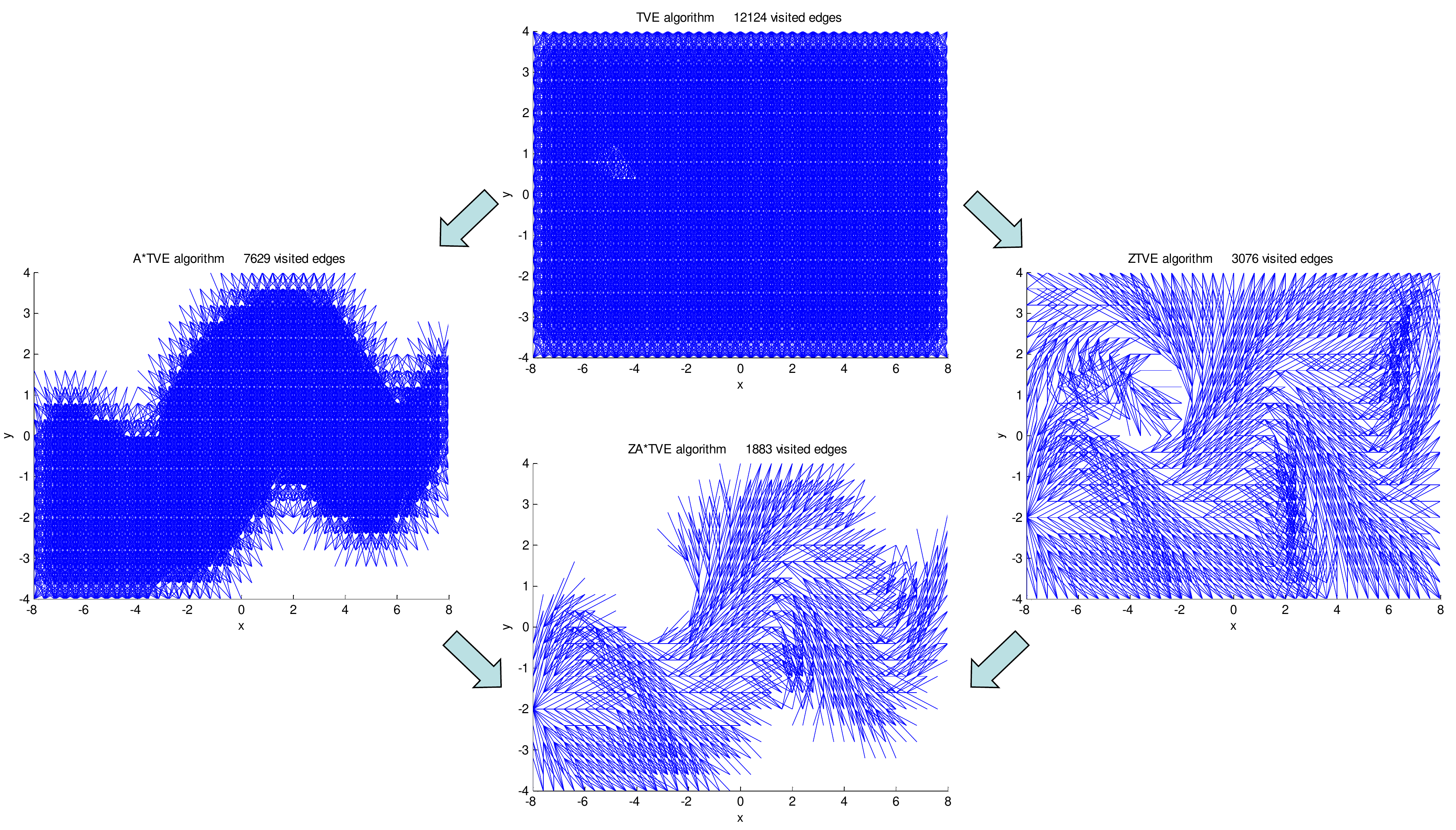}
	 \vspace{-20pt}
	 \caption{Visited edges using the several methods}
   \label{img:CompareMethods}
   	\vspace{-5pt}
\end{center}		
\end{figure}
This is reasonable since the algorithm includes only a subset of the vertices in the path search, in fact, only the preferred vertices with a short distance to the goal point. The inclusion of Zermelo's optimal navigation formula in the search algorithm (ZTVE) (see Section \ref{subsec:Zermelo} results in a decrease of the number of cost function calls to about one quarter of the calls using the ITVE algorithm. With both methods used together (ZA*TVE), the two merged acceleration mechanisms provide a further decrease of the number of cost function and current model calls. The use of the ZA*TVE algorithm allows a decrease of the number of cost function calls (CFC) by about a factor of 5, and, by a factor of 9 for the current model calls (CMC) in comparison to the  TVE algorithm. This improvement makes the practical use of the ZA*TVE algorithm possible for the case of (i) the computationally-intensive ocean current calculations, or, (ii) to determine the optimal departure time. \autoref{img:CompareMethods} shows the visited edges (blue lines) using the several search methods started from start position SP1.
\subsection{Possible missions and path smoothing algorithm}\label{sub:Missions}
\begin{table*}[!t]
	\centering
   \caption{Results of the path smoothing algorithm by using different missions}
	 \includegraphics[width=0.95\textwidth]{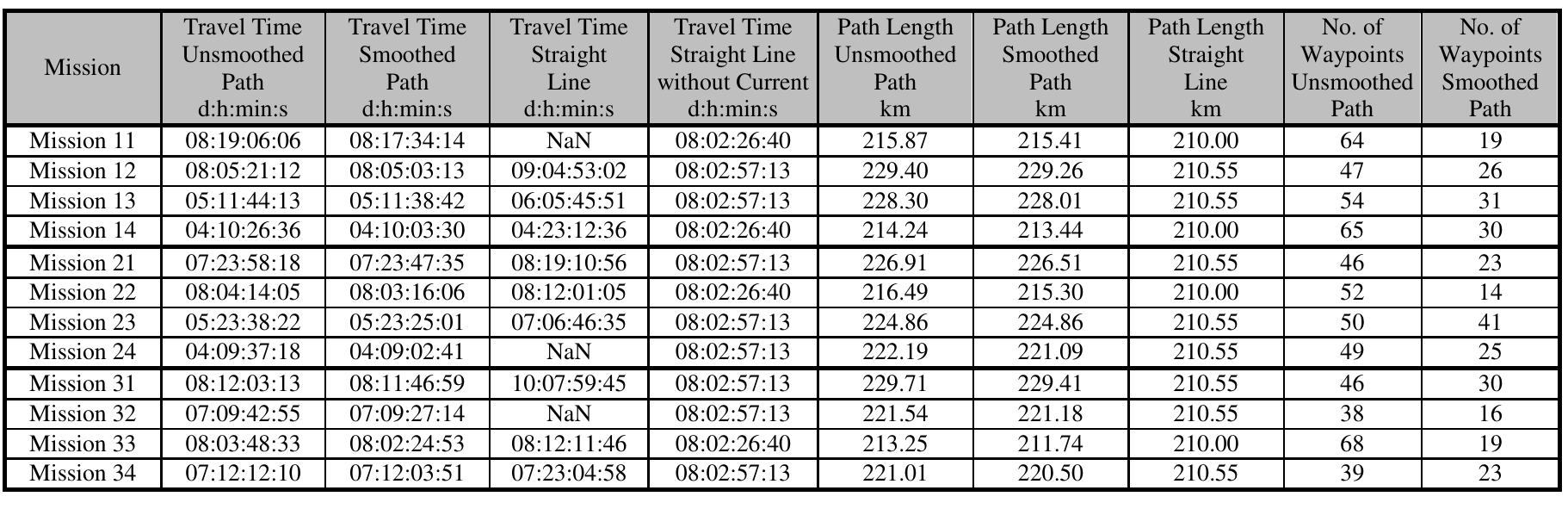}
   \label{tab:MissionsResults}
   \vspace{-15pt}
\end{table*}
This section presents the results using the path smoothing algorithm presented in Section \ref{sec:PathSmooting} by means of a selection of the possible missions along the Newfoundland and Labrador Shelf. \autoref{tab:MissionsResults} includes the results of the travel time and the length of the paths found, the smoothed path, and straight line to the goal point. 
Furthermore, the number of waypoints for the generated (unsmoothed) path and the smoothed path are shown. \autoref{img:Missions} shows the mission paths. The trajectories of the unsmoothed and the smoothed path of the missions are similar, so that the two lines are superimposed. 
\pagebreak
The length of the straight line for all missions is 210 km. The utilization of the Labrador Stream in Mission M13, M14, M23 and M24 brings a remarkable decrease of the mission endurance in comparison to the other missions. The number of waypoints in the missions can be decreased on average more than half using the smoothing algorithm, which improves the resulting path with respect to travel time. This is possible because new connections (edges) will be created which were not available in the geometrical graph during the search. An additional decrease of the waypoint list is possible,
\begin{figure}[hb]
	\centering
	\vspace{-5pt}
	\includegraphics[width=0.98\columnwidth]{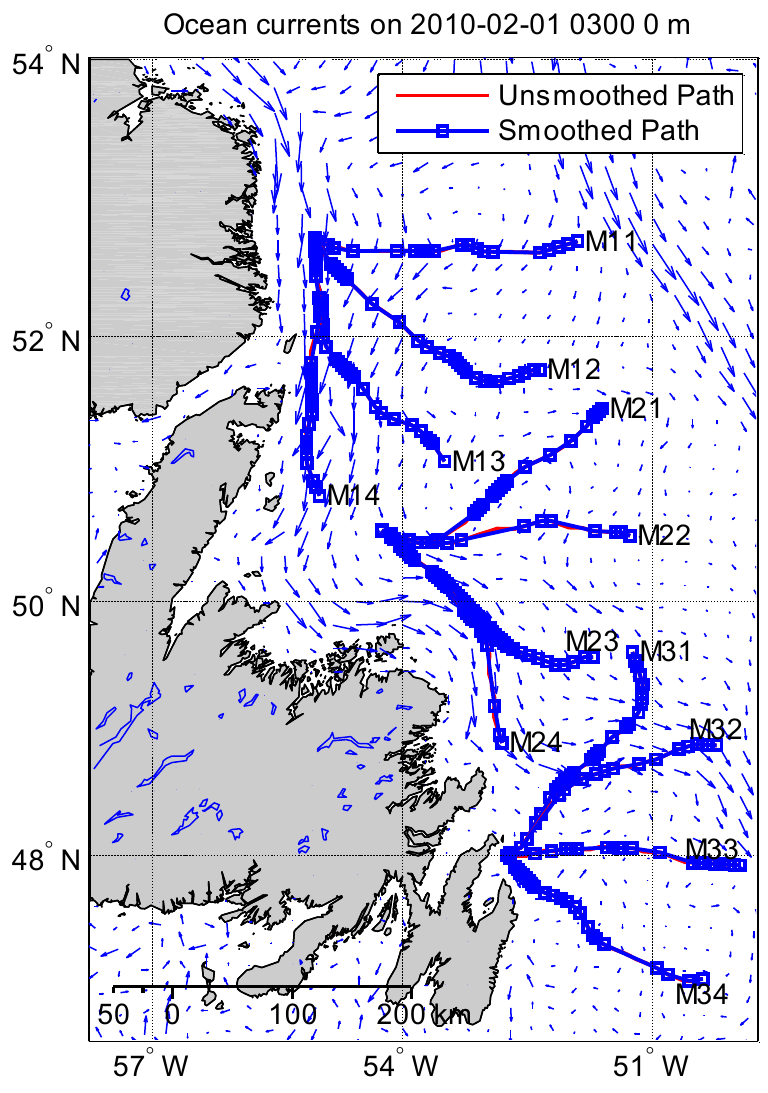}
	\vspace{-10pt}
	\caption{Time optimal path for different missions along the Newfoundland and Labrador Shelf}
	\label{img:Missions}
\end{figure}
when longer travel times to the goal point are accepted (see \autoref{tab:PathSmoothing}, Marker V). The direct course to the goal point leads to longer travel times or is impassable in the case of an adverse ocean current.
\section{Conclusion and future work}\label{sec:Conclusion}
In this paper algorithms for path planning in a time-varying environment based on graph methods are presented. Using the ocean current information in a geometrical graph, the position of the vertices and their possible connections (edges) are very important. This choice should consider the trend of the current flow and the possibility of optimal connections from one vertex to another in a given current field. Methods to accelerate the processing time of the basic TVE algorithm are described in the first part of the paper. The algorithms of the cost function for every connection are presented in the middle part of this paper. This requires the use of a fast calculation for the precise travel time from one vertex to another. In the last part of this paper, an algorithm to detect the optimal departure time is described. Current and future research topics are the analyses of possibilities for parallelization and the inclusion of inaccuracies in path planning as a result of forecast error variance, accuracy of calculation in the cost functions and a different observed vehicle speed in the real mission than planned \cite{Eichhorn2011a}. The presented path planning algorithms are aimed at saving time. It is also possible, however, to include the energy consumption in the cost function as shown in \cite{Alvarez2004,Zhang2008}. An additional research topic is the inclusion of a glider model which simulates the energy consumption in a glider \cite{Woithe2010}, to extract energy information for the cost function.
\section*{Acknowledgments}\label{sec:acknowledgments}
This work was financed by the German Research Foundation (DFG) within the scope of a two-year research fellowship (DFG-Number: EI 813/1-1). I would like to thank the National Research Council Canada Institute for Ocean Technology and in particular Dr. Christopher D. Williams for support during this project.





\clearpage 
\cleardoublepage 

\bibliographystyle{elsarticle-num}





\end{document}